\documentclass[journal]{IEEEtran}
\usepackage[switch]{lineno}
\usepackage{kotex}
\let\linenumbers\nolinenumbers\nolinenumbers

\usepackage{cite}
\usepackage[utf8]{inputenc} %
\usepackage[T1]{fontenc}    %
\usepackage{hyperref}
\usepackage{url}
\usepackage{booktabs}       %
\usepackage{amsfonts}       %
\usepackage{nicefrac}       %
\usepackage{microtype}      %
\usepackage{xcolor}
\usepackage{graphicx}
\usepackage{wrapfig}
\usepackage{enumitem}
\usepackage{comment}
\usepackage{tipa}
\usepackage{amsmath}
\usepackage{listings}
\usepackage{selectp}
\usepackage{multirow}
\usepackage[frozencache]{minted}
\usepackage{bbm}
\usepackage{bm}
\usepackage{amssymb}
\usepackage{graphicx}
\usepackage{subfig}
\usepackage{mathptmx}
\usepackage{adjustbox}

\definecolor{blue-violet}{rgb}{0.54, 0.17, 0.89}

\ifCLASSINFOpdf
\else
\fi

\begin{document}
\linenumbers
\title{
Pushing the Limits of Vision-Language Models

in Remote Sensing without Human Annotations}

\author{Keumgang~Cha,
        Donggeun Yu,
        Junghoon Seo%
\thanks{Keumgang Cha, Donggeun Yu, and Junghoon Seo are with SI Analytics, Daejeon 34051, South Korea (e-mail: chagmgang@si-analytics.ai; donggeun@si-analytics.ai; jhseo@si-analytics.ai).}%
}

\markboth{Pre-Print}%
{Shell \MakeLowercase{\textit{et al.}}: Bare Demo of IEEEtran.cls for Journals}

\maketitle

\begin{abstract}
The prominence of generalized foundation models in vision-language integration has witnessed a surge, given their multifarious applications. Within the natural domain, the procurement of vision-language datasets to construct these foundation models is facilitated by their abundant availability and the ease of web crawling. Conversely, in the remote sensing domain, although vision-language datasets exist, their volume is suboptimal for constructing robust foundation models. This study introduces an approach to curate vision-language datasets by employing an image decoding machine learning model, negating the need for human-annotated labels. Utilizing this methodology, we amassed approximately 9.6 million vision-language paired datasets in VHR imagery. The resultant model outperformed counterparts that did not leverage publicly available vision-language datasets, particularly in downstream tasks such as zero-shot classification, semantic localization, and image-text retrieval. Moreover, in tasks exclusively employing vision encoders, such as linear probing and k-NN classification, our model demonstrated superior efficacy compared to those relying on domain-specific vision-language datasets.
\end{abstract}


\IEEEpeerreviewmaketitle

\vspace{-2mm}
\section{Introduction}
\IEEEPARstart{F}oundation models are at the forefront of breakthrough in the deep learning community. Unlike specialized models that demand new labeling and training for different target tasks, foundation models boast of a flexible architecture that can efficiently span diverse tasks. This includes zero-shot classification, semantic localization, and even cross-modal retrieval. In the world of computer vision, seminal contributions like DINO\cite{caron2021emerging} and SAM\cite{kirillov2023segment} have carved a niche. Concurrently, the natural language processing domain has been revolutionized by models such as BERT\cite{devlin2018bert}, GPT3\cite{brown2020language}, and PaLM\cite{chowdhery2022palm}. Further amalgamating vision and language has led to transformative works such as Flamingo\cite{alayrac2022flamingo}, InstructBLIP\cite{instructblip}, and BEiT-3\cite{wang2022image}.

The remote sensing community, recognizing the potential of these models, has increasingly incorporated foundation models into its fold. Several works, prominently involving the Masked Image Modeling (MIM) approach, have made significant strides in tasks specific to this domain\cite{reed2022scale, wang2022advancing}. However, these models often encounter hurdles. A persistent challenge lies in their reliance on supervised fine-tuning, especially when deployed for core computer vision tasks.

Addressing these challenges has led to an intensified focus on vision-language foundation models within the remote sensing community. Specifically, the principles of contrastive learning between vision and language, exemplified by models like CLIP\cite{radford2021learning}, have gained traction. The allure of these models is their ability to adeptly manage a gamut of applications, often bypassing the tedious fine-tuning phase.

The bedrock of successful foundation models invariably remains quality datasets. Within the remote sensing context, although datasets like RSICD\cite{lu2017exploring} and UCM\cite{qu2016deep} exist, they often pale in comparison to voluminous datasets from more natural domains, such as LAION-5B\cite{schuhmann2022laion}. Methods to bridge this gap have been devised. For instance, RS5M\cite{zhang2023rs5m} employed the BLIP-2\cite{li2023blip} model to curate vision-language pairs, while RemoteCLIP\cite{liu2023remoteclip} aimed to convert traditional datasets into the vision-language format.

In this context, contribution of this paper is twofold: Firstly, we delineate a methodology to create a robust vision-language dataset tailored specifically for the remote sensing domain. By leveraging the potential of InstructBLIP\cite{instructblip}, we strive to ensure linguistic diversity and quality, sourcing images exclusively from esteemed remote sensing repositories. Secondly, building upon our crafted dataset, we introduce RSCLIP. Trained within the well-established CLIP framework\cite{radford2021learning}, RSCLIP promises to bridge the performance gap, outdoing models trained on synthetic labels and standing toe-to-toe with those reliant on human-annotated labels.

\vspace{-2mm}
\section{Proposed Method}

\subsection{Generation of Large-Scale Vision-Language Datasets}

The InstructBLIP\cite{instructblip} is utilized to extract vision-language pairs from individual images. Since InstructBLIP is tailored to echo the user's intent in generating captions, two distinct captions are produced for each image in this study. To guide the description of each image, the prompts "Write a short description for the image." and "Describe the image in detail" are provided, aiming to yield both concise and extended captions, respectively.

The source datasets employed to generate the vision-language pairs include fMoW\cite{christie2018functional}, Million-AID\cite{long2021creating}, DFC2019\cite{lian2020large}, DFC2021\cite{malkin2021high}, DeepGlobe\cite{demir2018deepglobe}, DIOR\cite{li2020object}, HRSC\cite{liu2017high}, and Inria\cite{maggiori2017can}. Given that the images sourced from these datasets vary in size, they are resized and cropped to a uniform 512 pixel square before being inputted into InstructBLIP. Additionally, subsets from RS5M, fMoW, and Million-AID are harnessed to pretrain the foundational model. In total, this process results in 9,686,720 vision-language pairs.

\vspace{-3mm}
\subsection{Dataset Statistics}

\begin{figure}[t!]
 \centering
 \includegraphics[width=0.28\textwidth]{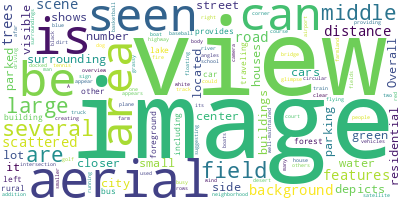}
 \includegraphics[width=0.20\textwidth]{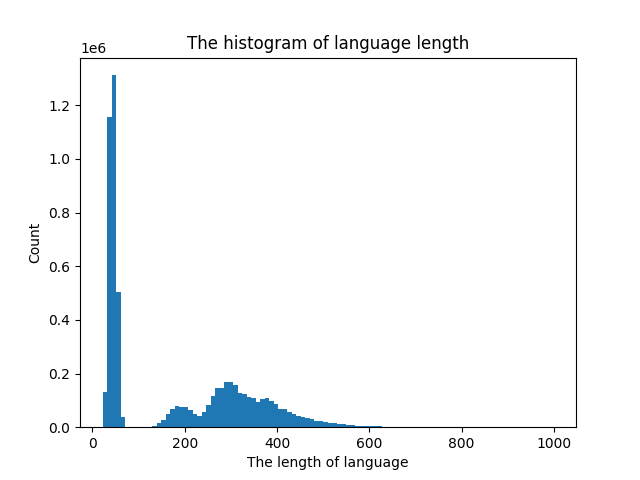}
 \vspace{-5mm}
 \caption{The word cloud and length of language generated by InstructBLIP.}
 \vspace{-6mm}
 \label{fig:statistics}
\end{figure}

Figure \ref{fig:statistics} presents both a word cloud and a histogram representing the distribution of the generated language. The vision-language data extracted from RS5M is excluded from this visualization. In the word cloud, we exclude determiners, prepositions, conjunctions, WH-pronouns, existentials, and adverbs, as these are primarily function words that serve to structure sentences rather than convey specific content. The histogram reveals two predominant distributions centered around a length of 100 words. This bifurcation results from our use of InstructBLIP to generate diverse language samples: descriptions below 100 words in length were produced in response to the prompt "Write a short description for the image.", while those exceeding 100 words were elicited with "Describe the image in detail." The total number of vision-language pairs employed to construct the RSCLIP is 9,686,720 of which 6,278,368 were generated by InstructBLIP and 3,408,352 were sourced from RS5M.

\vspace{-3mm}
\subsection{Pretraining Vision-Language Model}

CLIP models \cite{radford2021learning} are designed to ensure that items with similar meanings are located close in their representation space, while those with distinct meanings are positioned farther apart. The optimization of the CLIP model is achieved using the InfoNCE loss\cite{oord2018representation}. The CLIP model comprises a vision encoder and a text encoder. In this study, the vision encoder is based on the vision transformer model\cite{dosovitskiy2020image} with parameters set to: 16 patch size, 768 hidden size, 3072 MLP size, 12 heads, and 12 layers. The text encoder is the BERT-base model\cite{devlin2018bert} with a configuration of 768 hidden size, 3072 MLP size, 12 heads, and 12 layers. Notably, both the vision and text encoders are pre-trained using the Masked AutoEncoder on the Million-AID dataset\cite{wang2022advancing} and BERT-base, rather than being trained from scratch.

For pretraining, simple data augmentation is implemented. As InstructBLIP provides descriptions pertaining to the direction, position, and color of objects, strong augmentations such as aggressive resized random crops, random rotations, flips, and color distortions can introduce inconsistencies between the vision and language representations. Consequently, only resized random cropping ranging from 0.8 to 1.0 is employed. The input image size is set to 448. During pretraining, the batch size per GPU is 112, distributed across 16 GPUs for 10 epochs. The temperature parameter is set to 0.07. For optimization purposes, the base learning rate is 5.0e-4 / 32768, with a weight decay of 0.01. Therefore, the effective learning rate applied is 16 $\times$ 112 $\times$ 5.0e-4 / 32768. The optimization employs the AdamW optimizer, paired with a cosine decay scheduler and a single warm-up epoch.

\vspace{-2mm}
\section{Experiment}
We present results from both main and additional experiments across various downstream tasks. 
In all tables of exeperiment results, the best performance value in each column is bold and italicized.
 The main experiment results encompass image-text retrieval, zero-shot classification and semantic localization. Meanwhile, the additional experiment results include image-text retrieval, zero-shot classification, full-shot linear probing, k-NN classification and few-shot classification. The image-text retrieval, zero-shot classification and semantic localization are downstream tasks to evaluate the ability of cross modality. The full-shot linear probing, k-NN classification and few-shot classification are adopted to measure the uni modality of vision. The A distinguishing criterion between the main and additional experiments is whethe the compared models were using the downstream task datasets during pretraining. Specifically, the main experiment does not utilize the vision-language pairs from the downstream task, while the additional experiment does.
\vspace{-3mm}

\subsection{Main Experiment Results}

\begin{table*}[ht]
    {\scriptsize
    \renewcommand{\arraystretch}{1.2}
    \centering
    \begin{adjustbox}{width=0.9\textwidth}
    \begin{tabular}{cc ccc ccc c ccc ccc c}
    \hline
    \multicolumn{2}{c}{} & \multicolumn{7}{c}{RSICD} & \multicolumn{7}{c}{RSITMD}\\
    \multicolumn{2}{c}{} & \multicolumn{3}{c}{Image-to-Text} & \multicolumn{3}{c}{Text-to-Image} & & \multicolumn{3}{c}{Image-to-Text} & \multicolumn{3}{c}{Text-to-Image} & \\
    Model & Params & $R@1$ & $R@5$ & $R@10$ & $R@1$ & $R@5$ & $R@10$ & mR & $R@1$ & $R@5$ & $R@10$ & $R@1$ & $R@5$ & $R@10$ & mR \\
    \hline
    CLIP(ViT-B-32)\cite{zhang2023rs5m} & $\approx 151M$ & 5.4 & 15 & 24.06 & 6.44 & 19.82 & 30.28 & 16.83 & 9.51 & 23.01 & 32.74 & 8.81 & 27.92 & 43.23 & 24.20 \\
    CLIP(ViT-L-14)\cite{zhang2023rs5m} & $\approx 427M$ & - & - & - & - & - & - & - & 12.61 & 29.87 & 42.48 & \textit{\textbf{15.17}} & 39.2 & 52.92 & 32.04 \\
    CLIP(ViT-H-14)\cite{zhang2023rs5m} & $\approx 986M$ & - & - & - & - & - & - & - & 12.61 & 33.41 & 44.69 & 14.2 & 39.47 & 55.27 & 33.28 \\
    CLIP(ViT-bigG-14)\cite{zhang2023rs5m} & $\approx 2500M$ & - & - & - & - & - & - & - &13.94 &34.51 &45.13 &13.98 &41.59 &56.59 &34.29 \\
    VSE++\cite{faghri2017vse++} & - & 3.38 & 9.51 & 17.46 & 2.82 & 11.32 & 18.1 & 10.43 &10.38 &27.65 &39.6 &7.79 &24.87 &38.67 &24.83 \\
    AFMFN\cite{yuan2022exploring} & - & 5.39 & 15.08 & 23.4 & 4.9 & 18.28 & 31.44 & 16.42 &11.06 &29.2 &38.72 &9.96 &34.03 &52.96 &29.32 \\
    KCR\cite{mi2022knowledge} & - & 5.84 & 22.31 & 36.12 & 4.76 & 18.59 & 27.2 & 19.14 & - & - & - & - & - & - & - \\
    GaLR\cite{yuan2022remote} & - & 6.59 & 19.85 & 31.04 & 4.69 & 19.48 & 32.13 & 18.96 &14.82 &31.64 &42.48 &11.15 &36.68 &51.68 &31.41\\
    Pfeiffer\cite{zhang2023rs5m} & $\approx 152M$ & 7.87 & 18.21 & 27.26 & 5.84 & 20.57 & 33.14 & 18.82 &11.5 &25 &36.28 &9.65 &31.59 &46.9 &26.82 \\
    Prefixtuning\cite{zhang2023rs5m} & $\approx 152M$ & 9.61 & 22.05 & 32.11 & 6.99 & 22.09 & 33.06 & 20.99 &13.72 &30.97 &43.14 &6.25 &30.04 &47.26 &28.56\\
    LoRA\cite{zhang2023rs5m} & $\approx 152M$ & 7.14 & 18.48 & 27.17 & 6.18 & 19.05 & 29.66 & 17.95 &13.5 &28.98 &39.38 &6.86 &26.55 &40.53 &25.97\\
    UniPELT\cite{zhang2023rs5m} & $\approx 152M$ & 8.87 & 21.04 & 31.29 & 6.81 & 24.01 & 35.75 & 21.30 &13.27 &29.2 &41.37 &9.69 &32.57 &48.36 &29.08\\
    RSCLIP & $\approx 197M$ & \textit{\textbf{10.43}} & \textit{\textbf{25.34}} & \textit{\textbf{39.34}} & \textit{\textbf{9.9}} & \textit{\textbf{30.52}} & \textit{\textbf{45.03}} & \textit{\textbf{26.76}} & \textit{\textbf{19.25}} & \textit{\textbf{36.06}} & \textit{\textbf{46.68}} &12.92 & \textit{\textbf{42.04}} & \textit{\textbf{63.14}} & \textit{\textbf{36.68}} \\
    \hline
    \end{tabular}
    \end{adjustbox}
    \vspace{-2mm}
    \caption{Image-text retrieval in both RSICD and RSITMD dataset. As main experiment results, the models included RS5M is used as comparison. The RSCLIP shows the highest performance in all data sets and all top-k except for R@1 in RSITMD.}
    \label{tab:retrieval_main}
    }
    \vspace{-5mm}
\end{table*}

\begin{table}[ht]
    \renewcommand{\arraystretch}{1.2}
    \centering
    \begin{adjustbox}{width=0.49\textwidth}
    \begin{tabular}{cc ccc cccc}
    \hline
    \multicolumn{2}{c}{} & \multicolumn{3}{c}{Zero-shot Classification} & \multicolumn{4}{c}{Semantic Localization}\\
    \multicolumn{2}{c}{} & AID & RESISC45 & Avg & \multicolumn{4}{c}{AIR-SLT}\\
    Model & Params & \multicolumn{3}{c}{Top-1 Accuracy} & $R_{su}$ $\uparrow$ & $R_{as}$ $\downarrow$ & $R_{da}$ $\uparrow$ & $R_{mi}$ $\uparrow$ \\
    \hline
    CLIP(ViT-B-32)\cite{zhang2023rs5m} & $ 151M$ & 60.84 & 58.97 & 59.91 & 0.7220 & \textit{\textbf{0.2848}} & 0.6880 & 0.7111 \\
    SeLov1\cite{yuan2022learning} & - & - & - & - & 0.6920 & 0.3323 & 0.6667 & 0.6772\\ 
    SeLov2\cite{yu2023selo} & - & - & - & - & 0.7199 & 0.2925 & 0.6658 & 0.7021\\ 
    Pfeffier\cite{zhang2023rs5m} & $ 152M$ & 68.37 & 67.79 & 68.08& 0.7180 & 0.3116 & 0.6589 & 0.6912\\
    Prefixtuning\cite{zhang2023rs5m} & $ 152M$ & 69.83 & 66.74 & 68.29& 0.7241 & 0.3132 & 0.6867 & 0.7017\\
    LoRA\cite{zhang2023rs5m} & $ 152M$ & 67.38 & 65.53 & 66.46& 0.7176 & 0.2857 & 0.6911 & 0.7098\\
    UniPELT\cite{zhang2023rs5m} & $ 152M$ & 70.92 & 66.61 & 68.77& 0.7292 & 0.3463 & 0.6461 & 0.6820\\
    RSCLIP & $ 192M$ & \textit{\textbf{75.82}} & \textit{\textbf{68.59}} & \textit{\textbf{72.20}} & \textit{\textbf{0.7349}} & 0.2877 & \textit{\textbf{0.7070}} & \textit{\textbf{0.7200}} \\
    \hline
    \end{tabular}
    \end{adjustbox}
    \vspace{-2mm}
    \caption{The zero-shot classification and semantic localization results. In zero-shot classification, the RSCLIP has the best performance as shown in table. In semantic localization, the RSCLIP records the best performance except for $R_{as}$. 
    }
    \label{tab:zero main}
    \vspace{-6mm}
\end{table}

\subsubsection{Image-Text Retrieval}

We assess RSCLIP's capabilities on two image-text retrieval benchmark datasets, RSICD and RSITMD. For this task, we extract test split datasets. Both images and texts serve as input for the respective encoders, undergoing L2 normalization. Post-normalization, representation similarities are gauged using dot-products – a standard similarity measurement technique. Retrieval metrics comprise retrieval recall for top-1 (R@1), top-5 (R@5), top-10 (R@10), and their mean recall. Table \ref{tab:retrieval_main} provides detailed image-text retrieval results. Across all datasets and top-k metrics except for R@1 in RSITMD, RSCLIP surpasses previous methods, displaying both the best individual and mean recall performances.

\subsubsection{Zero Shot Classification}

For evaluation on the zero-shot image classification, we employed two remote sensing scene classification datasets: AID\cite{xia2017aid} and RESISC45\cite{cheng2017remote}, with the latter representing a key VHR scene classification dataset. We applied the standard template-based prompting method, using "a satellite image of {class name}" to create the zero-shot classifier. Table \ref{tab:zero main} details the evaluation results for zero-shot classification. Within it, RSCLIP demonstrates superior accuracy on both AID and RESISC45 datasets, boasting the best performance across datasets and the highest average performance.

\subsubsection{Semantic Localization}

To gauge semantic localization in expansive remote sensing imagery, we used the AIR-SLT\cite{yuan2022learning} dataset. Metrics $R_{su}$, $R_{as}$, $R_{da}$, and $R_{mi}$ are reported in Table \ref{tab:zero main}. Here, $R_{su}$ denotes the proportion of significant areas, $R_{as}$ measures the deviation between the semantic localization map's probability center and the ground truth (GT) center, $R_{da}$ quantifies attention dispersion, and $R_{mi}$ is a mean indicator defined as $R_{mi}=w_{su}R_{su}+w_{as}(1-R_{as})+w_{da}R_{da}$. Higher values of $R_{su}$, $R_{da}$, and $R_{mi}$ are preferable, while a lower $R_{as}$ value is desirable. The evaluation metrics follow the original research's hyperparameters\cite{yuan2022learning}. 
Although the RSCLIP shows the superior performance except for $R_{as}$, the RSCLIP's $R_{mi}$, which is their comprehensive indicator, records the best performance.

\vspace{-3mm}
\subsection{Additional Experiment Results}

\begin{table*}[ht]
    {\scriptsize
    \renewcommand{\arraystretch}{1.2}
    \centering
    \begin{adjustbox}{width=1.0\textwidth}
    \begin{tabular}{cc cccc cccc cccc cccc }
    \hline
    \multicolumn{2}{c}{} & \multicolumn{4}{c}{RSICD} & \multicolumn{4}{c}{RSITMD} & \multicolumn{4}{c}{UCM} & \multicolumn{4}{c}{Sydney} \\
    \multicolumn{2}{c}{} & \multicolumn{2}{c}{Image-to-Text} & \multicolumn{2}{c}{Text-to-Image} & \multicolumn{2}{c}{Image-to-Text} & \multicolumn{2}{c}{Text-to-Image}& \multicolumn{2}{c}{Image-to-Text} & \multicolumn{2}{c}{Text-to-Image}& \multicolumn{2}{c}{Image-to-Text} & \multicolumn{2}{c}{Text-to-Image} \\
    Model & Params & $R@1$ & $R@5$ & $R@1$ & $R@5$ & $R@1$ & $R@5$ & $R@1$ & $R@5$ & $R@1$ & $R@5$ & $R@1$ & $R@5$ & $R@1$ & $R@5$ & $R@1$ & $R@5$ \\
    \hline
    S-CLIP($L=U$)$\lozenge$\cite{mo2023s} & $\approx 102M$ & 4.2 & 18.4 & 4.2 & 16.8 & - & - & - & - & 11.6 & 45.7 & 11.1 & 43.5 & 14.9 & 50 & 17.8 & 55.1 \\
    S-CLIP($L \neq U$)$\lozenge$\cite{mo2023s} & $\approx 102M$ & 4.2 & 17.1 & 3.9 & 15.8 & - & - & - & - & 9.8 & 43.5 & 10.8 & 42.5 & 13.8 & 48.9 & 17.8 & 52.3 \\
    RemoteCLIP$\lozenge$\cite{liu2023remoteclip} & $\approx 102M$ & 13.36 & 32.94 & 10.76 & 32.83 & 23.67 & 47.57 & 19.29 & 51.55 & 13.33 & 50.48 & 15.24 & 57.14 & - & - & - & - \\
    RemoteCLIP$\lozenge$\cite{liu2023remoteclip} & $\approx 151M$ & 17.02 & \textit{\textbf{37.97}} & 13.71 & 37.11 & 27.88 & 50.66 & 22.17 & 56.46 & \textit{\textbf{20.48}} & \textit{\textbf{59.85}} & \textit{\textbf{18.67}} & 61.52 & - & - & - & - \\
    RemoteCLIP$\lozenge$\cite{liu2023remoteclip} & $\approx 428M$ & \textit{\textbf{18.39}} & 37.42 & \textit{\textbf{14.73}} & \textit{\textbf{39.93}} & \textit{\textbf{28.76}} & \textit{\textbf{52.43}} & \textit{\textbf{23.76}} & \textit{\textbf{59.51}} & 19.05 & 54.29 & 17.71 & 62.19 & - & - & - & - \\
    RSCLIP$\blacklozenge$ & $\approx 197M$ & 10.43 & 25.34 & 9.9 & 30.52 & 19.25 & 36.06 & 12.92 & 42.04 & 19.05 & 56.19 & 16.38 & \textit{\textbf{62.29}} & \textit{\textbf{29.31}} & \textit{\textbf{58.62}} & \textit{\textbf{22.07}} & \textit{\textbf{57.93}} \\
    \hline
    \end{tabular}
    \end{adjustbox}
    \vspace{-2mm}
    \caption{The additional evaluation results of image-text retrieval in RSICD, RSITMD, UCM and Sydney dataset. In this experiment, although the RSCLIP is not trained with vision-language pairs presented in the downstream tasks, it can be seen in table that the RSCLIP shows the performance that is just as good as the model using it.}
    \label{tab:retrieval_addition}
    }
    \vspace{-3mm}
\end{table*}

\begin{table*}[ht]
    \setlength{\tabcolsep}{1pt}
    \renewcommand{\arraystretch}{1.2}
    \centering
    \scalebox{0.9}{
    \begin{tabular}{cc c c c c c c c c c c c c c c c c}
    \hline
     &  & RSICD-CLS & UCM-CLS & WHU-RS19 & AID & RESISC45 & EuroSAT & RSI-CB128 & RSI-CB256 & MLRSNet & PatternNet &  &  \\
    Method & Params & \multicolumn{10}{c}{Top-1 Accuracy} & Avg 1 & Avg 2 \\
    \hline
    S-CLIP(ResNet-50)$\Diamond$ \cite{mo2023s} & $\approx 102M$ &  66.90 & 66.70 & 86.90 & 73.00 & - & - & - & - & - & - & 73.38 \\
    S-CLIP(ViT-Base)$\Diamond$\cite{mo2023s} & $\approx 151M$ & \textit{\textbf{87.40}} & \textit{\textbf{88.90}} & \textit{\textbf{97.30}} & \textit{\textbf{93.10}} & - & - & - & - & - & - & \textit{\textbf{91.67}} & - \\
    RemoteCLIP(ResNet-50)$\Diamond$\cite{liu2023remoteclip} & $\approx 102M$ & - & - & 95.15 & 86.55 & 53.24 & 17.19 & 13.95 & 33.03 & 40.68 & 45.51 & - & 48.16 \\
    RemoteCLIP(ViT-Base)$\Diamond$\cite{liu2023remoteclip} & $\approx 151M$ & - & - & 96.12 & 91.30 & \textit{\textbf{70.33}} & 35.96 & 24.18 & 39.50 & 59.28 & 57.71 & - & 59.30 \\
    RSCLIP$\blacklozenge$ & $\approx 197M$ & 69.33 & 68.33 & 86.67 & 75.82 & 68.59 & \textit{\textbf{48.44}} & \textit{\textbf{30.59}} & \textit{\textbf{47.19}} & \textit{\textbf{65.12}} & \textit{\textbf{66.74}} & 75.04 & \textit{\textbf{61.14}} \\
    \hline
    \end{tabular}
    }
    \vspace{-2mm}
    \caption{The zero-shot classification with text prompt, which is "the satellite image of {class name}". The RSCLIP shows the competitive performance without using the vision-language pairs of the downstream tasks.}
    \label{tab:zero_addition}
    \vspace{-6mm}
\end{table*}

Distinct from the approach in this paper, we also contrasted RSCLIP with models like S-CLIP and RemoteCLIP, which directly utilize vision-language pairs. S-CLIP employs a semi-supervised technique, capitalizing on only 10\% of existing vision-language pairs. However, because its text encoder was informed directly by the vision-language pair, it's classified as an additional experiment. Similarly, RemoteCLIP, which learned all vision-language pairs directly, was also placed in this category.

Generally, RSCLIP doesn't top the charts in downstream tasks. This is expected as other models benefit from text encoders directly trained on downstream language distributions. Yet, RSCLIP remains competitive even without this advantage. Impressively, in tasks like few-shot, linear probing, and k-NN Classification, RSCLIP reigns supreme using only a vision encoder. For clarity, in the Additional Experiment Results section, models directly leveraging vision-language pairs are marked with $\Diamond$, while those that didn't utilize them at all bear the $\blacklozenge$ symbol. Detailed results follow below.

\subsubsection{Image-Text Retrieval}
For evaluation metric in retrieval, the retrieval recall of top-1 (R@1), and top-5 (R@5) are reported. Table \ref{tab:retrieval_addition} displays image-text retrieval results. Expectedly, RemoteCLIP, trained on the most direct vision-language pairs, outshines the rest. Still, when compared to S-CLIP, RSCLIP displays superior performance even without the direct 10\% vision-language advantage. This indicates the potential of our vision-language pair generation method.

\subsubsection{Zero Shot Classification}
Table \ref{tab:zero_addition} presents the top-1 accuracy for zero-shot classification across multiple datasets. For this evaluation, we utilized ten downstream datasets, including RSICD-CLS, UCMerced Land Use (UCM-CLS)\cite{yang2010bag}, WHU-RS19\cite{xia2009structural}, AID\cite{xia2017aid}, RESISC45\cite{cheng2017remote}, EuroSAT\cite{helber2019eurosat}, RSI-CB128\cite{li2020rsi}, RSI-CB256\cite{li2020rsi}, MLRSNet\cite{qi2020mlrsnet}, and PatternNet\cite{zhou2018patternnet}. Within the table, "Avg 1" represents the average performance across RSICD-CLS, UCM-CLS, WHU-RS19, and AID datasets and serves as a comparison with S-CLIP. "Avg 2" calculates the average for datasets WHU-RS19, AID, RESISC45, EuroSAT, RSI-CB128, RSI-CB256, MLRSNet, and PatternNet, intended for comparison with RemoteCLIP.

Regarding "Avg 1", RSCLIP, despite not immediately employing language, displays accuracy surpassing the ResNet-50 variant of S-CLIP, yet falling short of its ViT-Base counterpart. In the "Avg 2" category, RSCLIP doesn't top the charts for WHU-RS19, AID, and RESISC45. However, it excels in RSI-CB128, RSI-CB256, MLRSNet, and PatternNet. Moreover, in terms of average performance, RSCLIP achieves the highest score. Collectively, while it seems optimal to directly incorporate vision-language from the downstream task, our method of constructing a vision-language pair yields comparable results.

\subsubsection{Few-shot Classification}

Few-shot classification evaluates the standalone vision encoder. Datasets are split into training and testing sets at a ratio of 0.8 to 0.2. Depending on the settings, images from the training set are extracted per class based on the designated number of shots. These extracted images provide representations for shots, serving as training features for the linear probing model. Upon training this model, test images are transformed into representations via the vision encoder, then input into the trained model to predict image classes. For this experiment, datasets RSI-CB128, RSI-CB256, EuroSAT, MLRSNet, PatternNet, RESISC45, AID, and WHU-RS19 were employed. Shot numbers for few-shot classification were set at 1, 4, 8, 16, and 32, with the logistic regression model from scikit-learn functioning as the linear probing model. Table \ref{tab:few_shot} indicates that, despite RSCLIP not using direct vision-language pairs from the downstream task dataset, it surpasses RemoteCLIP across all few-shot settings, even in average accuracy only except for 1-shot classification in RESISC45. Two potential reasons underpin this outcome. Firstly, only the vision encoder is deployed in few-shot classification. Secondly, RSCLIP's pretraining phase utilized a significantly larger image corpus than RemoteCLIP.

\subsubsection{Full-shot Linear Probing and k-NN Classification}

Full-shot linear probing can be viewed as an extension of the few-shot classification. Unlike its few-shot counterpart where a limited number of images serve as input features for the linear probing model, full-shot classification utilizes all training-split images for this purpose. For k-NN classification, the k parameter for nearest neighbors is consistently set to 20, aligning with RemoteCLIP's approach\cite{liu2023remoteclip}. FAISS\cite{johnson2019billion} underpins the k-NN algorithm. Datasets RSI-CB128, RSI-CB256, EuroSAT, MLRSNet, PatternNet, RESISC45, AID, and WHU-RS19 were harnessed as benchmark datasets for these evaluations. Except for four cases, the table \ref{tab:lin_knn_addition} reveals that the RSCLIP consistently outperforms other models in both full-shot linear probing and k-NN classification across all datasets. The four cases includes both the linear probing of EuroSAT, RESISC45 and k-NN classification in RSI-CB128, RSI-CB256. However, its average performance also stands unmatched. These outcomes might stem from reasons similar to those discussed in the few-shot classification section.

\begin{table*}[ht]
    \setlength{\tabcolsep}{3pt}
    \renewcommand{\arraystretch}{1.2}
    \centering
    \scalebox{0.8}{
    \begin{tabular}{ccc c c c c c c c c c c}
    \hline
    Method & Backbone & Shot & RSI-CB128 & RSI-CB256 & EuroSAT & MLRSNet & PatternNet & RESISC45 & AID & WHU-RS19 & Avg \\
    \hline
    RemoteCLIP$\Diamond$ & ResNet-50 & \multirow{3}{*}{1} & 35.59 & 42.52 & 43.20 & 31.75 & 46.10 & 39.33 & 36.95 & 45.15 & 40.07 \\
    RemoteCLIP$\Diamond$ & ViT-Base & & 34.31 & 44.28 & 44.89 & 34.14 & 45.98 & \textit{\textbf{42.10}} & 37.04 & 40.78 & 40.44 \\
    RSCLIP$\blacklozenge$ & ViT-Base & & \textit{\textbf{60.65}} & \textit{\textbf{83.28}} & \textit{\textbf{54.13}} & \textit{\textbf{78.44}} & \textit{\textbf{82.38}} & 37.37 & \textit{\textbf{97.62}} & \textit{\textbf{100.00}} & \textit{\textbf{74.23}} \\
    \hline
    RemoteCLIP$\Diamond$ & ResNet-50 & \multirow{3}{*}{4} & 60.04 & 65.44 & 55.53 & 46.90 & 66.99 & 52.11 & 63.13 & 73.59 & 60.47 \\
    RemoteCLIP$\Diamond$ & ViT-Base &  &64.49 & 70.33 & 55.99 & 54.52 & 70.98 & 60.91 & 65.59 & 68.16 & 63.87 \\
    RSCLIP$\blacklozenge$ & ViT-Base &  & \textit{\textbf{80.65}} & \textit{\textbf{88.66}} & \textit{\textbf{73.53}} & \textit{\textbf{96.00}} & \textit{\textbf{98.25}} & \textit{\textbf{78.00}} & \textit{\textbf{99.52}} & \textit{\textbf{100.00}} & \textit{\textbf{89.33}} \\
    \hline
    RemoteCLIP$\Diamond$ & ResNet-50 & \multirow{3}{*}{8} &69.55 & 75.89 & 61.75 & 55.02 & 77.07 & 61.75 & 70.50 & 85.44 & 69.62 \\
    RemoteCLIP$\Diamond$ & ViT-Base &  &76.13 & 83.73 & 65.76 & 64.24 & 82.53 & 70.92 & 75.72 & 80.68 & 74.96 \\
    RSCLIP$\blacklozenge$ & ViT-Base &  &\textit{\textbf{89.35}} & \textit{\textbf{96.72}} & \textit{\textbf{75.50}} & \textit{\textbf{95.51}} & \textit{\textbf{99.00}} & \textit{\textbf{88.71}} & \textit{\textbf{98.57}} & \textit{\textbf{100.00}} & \textit{\textbf{92.92}} \\
    \hline
    RemoteCLIP$\Diamond$ & ResNet-50 & \multirow{3}{*}{16} &77.58 & 83.72 & 70.36 & 59.74 & 82.93 & 69.51 & 75.12 & 89.32 & 76.04 \\
    RemoteCLIP$\Diamond$ & ViT-Base &  &82.63 & 89.12 & 75.73 & 67.45 & 88.13 & 75.83 & 81.05 & 89.51 & 81.18 \\
    RSCLIP$\blacklozenge$ & ViT-Base &  &\textit{\textbf{94.84}} & \textit{\textbf{98.21}} & \textit{\textbf{94.10}} & \textit{\textbf{96.34}} & \textit{\textbf{99.00}} & \textit{\textbf{88.29}} & \textit{\textbf{99.05}} & \textit{\textbf{100.00}} & \textit{\textbf{96.23}} \\
    \hline
    RemoteCLIP$\Diamond$ & ResNet-50 & \multirow{3}{*}{32} &82.02 & 87.04 & 77.44 & 64.99 & 88.32 & 75.71 & 82.46 & 93.79 & 81.47 \\
    RemoteCLIP$\Diamond$ & ViT-Base &  &88.11 & 91.83 & 83.30 & 71.58 & 91.87 & 81.77 & 86.67 & 93.40 & 86.07 \\
    RSCLIP$\blacklozenge$ & ViT-Base &  &\textit{\textbf{96.77}} & \textit{\textbf{99.10}} & \textit{\textbf{95.60}} & \textit{\textbf{97.12}} & \textit{\textbf{99.63}} & \textit{\textbf{88.43}} & \textit{\textbf{98.81}} & \textit{\textbf{100.00}} & \textit{\textbf{96.93}} \\
    \hline
    \end{tabular}
    }
    \vspace{-2mm}
    \caption{The few-shot classification results in additional experiment. The RSCLIP is compared with the RemoteCLIP in various scene classification dataset. In all datasets and all k-shot settings, the RSCLIP is the best performance with the same reason of full linear probing and k-NN classification.}
    \label{tab:few_shot}
    \vspace{-3mm}
\end{table*}

\begin{table*}[ht]
    \setlength{\tabcolsep}{3pt}
    \renewcommand{\arraystretch}{1.2}
    \centering
    \scalebox{0.8}{
    \begin{tabular}{cc cc cc cc cc cc cc cc cc cc}
    \hline
    \multicolumn{2}{c}{} & \multicolumn{2}{c}{RSI-CB128} & \multicolumn{2}{c}{RSI-CB256} & \multicolumn{2}{c}{EuroSAT} & \multicolumn{2}{c}{MLRSNet} & \multicolumn{2}{c}{PatternNet} & \multicolumn{2}{c}{RESISC45} & \multicolumn{2}{c}{AID} & \multicolumn{2}{c}{WHU-RS19} & \multicolumn{2}{c}{Avg} \\
    \hline
    Method & Backbone & Linear & k-NN & Linear & k-NN & Linear & k-NN & Linear & k-NN & Linear & k-NN & Linear & k-NN & Linear & k-NN & Linear & k-NN & Linear & k-NN \\
    \hline
ImageNet$\blacklozenge$     & ResNet-50   & 95.69  & 93.24  & 97.92  & 97.40  & 91.48  & 88.41  & 78.98  & 74.78  & 96.18  & 93.45   & 86.16   & 83.60   & 83.00   & 79.45   & 95.63   & 90.21   & 90.63   & 87.57  \\
 SwAV$\blacklozenge$         & ResNet-50   & 95.27  & 95.61  & 98.29  & 98.17  & 91.17  & 91.37  & 79.04  & 76.12  & 96.94  & 94.18   & 88.60   & 85.59   & 86.00   & 80.80   & 96.12   & 92.23   & 91.43   & 89.26   \\
 Barlow Twins$\blacklozenge$ & ResNet-50   & 98.07  & 95.91  & 99.03  & 98.13  & 94.78  & 91.57  & 82.41  & 77.55  & 97.73  & 93.83   & 91.10   & 86.10   & 88.25   & 81.75   & 97.09   & 91.75   & 93.56   & 89.57\\   
 VICReg$\blacklozenge$       & ResNet-50   & 97.47  & 96.03  & 98.67  & 98.21  & 95.06  & 91.44  & 82.59  & 78.02  & 98.83  & 94.03   & 91.03   & 86.75   & 88.10   & 81.50   & 96.60   & 90.78   & 93.54   & 89.60  \\ 
 CLIP$\blacklozenge$         & ResNet-50   & 94.89  & \textit{\textbf{97.05}}  & 97.30  & 97.24  & 91.67  & 88.54  & 80.08  & 77.14  & 95.61  & 92.86   & 85.73   & 85.65   & 90.95   & 86.90   & 97.57   & 93.69   & 91.73   & 89.88   \\
 CLIP-CL$\lozenge$      & ResNet-50   & 95.99  & 94.92  & 98.41  & 98.09  & 89.80  & 87.65  & 79.32  & 76.99  & 97.30  & 95.15   & 89.10   & 88.19   & 94.80   & 92.85   & 98.06   & 97.57   & 92.85   & 91.43 \\  
 ImageNet$\blacklozenge$     & ViT-Base    & 96.45  & 91.29  & 98.11  & 97.00  & 85.57  & 76.56  & 78.61  & 74.05  & 96.81  & 92.98   & 86.89   & 81.63   & 83.55   & 76.45   & 94.17   & 89.81   & 90.02   & 84.97   \\
 ViTAE$\blacklozenge$        & ViT-Base    & 93.10  & 95.65  & 98.41  & 94.05  & 61.41  & 82.27  & 91.15  & 80.37  & 98.50  & 90.82   & 87.94   & 65.33   & 88.30   & 64.05   & 91.74   & 70.39   & 88.82   & 80.37   \\
 CLIP$\blacklozenge$         & ViT-Base    & 97.36  & 94.17  & 98.55  & 97.40  & 95.15  & 90.28  & 85.43  & 82.26  & 97.58  & 94.36   & 92.60   & 89.73   & 94.95   & 90.35   & 97.09   & 93.69   & 94.84   & 91.53  \\
 RemoteCLIP$\lozenge$   & ResNet-50   & 96.06  & 94.78  & 98.39  & 97.62  & 92.56  & 90.20  & 83.32  & 81.21  & 97.37  & 95.95   & 90.94   & 90.05   & 94.35   & 92.10   & 98.06   & 95.63   & 93.88   & 92.19   \\
 RemoteCLIP$\lozenge$   & ViT-Base    & 98.02  & 95.82  & 99.01  & \textit{\textbf{98.51}}  & \textit{\textbf{96.19}}  & 93.50  & 87.00  & 85.11  & 98.47  & 97.32   & \textit{\textbf{94.27}}   & 92.67   & 95.95   & 92.55   & 97.57   & 74.17   & 95.81   & 91.21   \\
 RSCLIP$\blacklozenge$       & ViT-Base    & \textit{\textbf{98.13}}  & 96.70  & \textit{\textbf{99.09}}  & 98.02  & 95.50  & \textit{\textbf{94.33}}  & \textit{\textbf{94.01}}  & \textit{\textbf{93.36}}  & \textit{\textbf{99.08}}  & \textit{\textbf{98.60}}   & 94.14   & \textit{\textbf{93.64}}   & \textit{\textbf{97.95}}   & \textit{\textbf{97.65}}   & \textit{\textbf{99.50}} & \textit{\textbf{98.01}} & \textit{\textbf{97.18}} & \textit{\textbf{96.29}}\\
    \hline
    \end{tabular}
    }
    \vspace{-2mm}
    \caption{The full linear probing and k-NN classification in additional experiment. As mentioned, the $\lozenge$ is the model trained with direct vision-language pair of downstream tasks and the $\blacklozenge$ is the model not using the direct language expression of downstream tasks. Although the RSCLIP is marked as $\blacklozenge$, the RSCLIP scores the best performance in all dataset because this downstream tasks require only vision encoder.}
    \label{tab:lin_knn_addition}
    \vspace{-6mm}
\end{table*}

\vspace{-2mm}
\section{Conclusion}
This paper demonstrates the potential of leveraging large language models for image decoding to construct vision-language models without the need for human-annotated labels. We introduced a vision-language foundational model, RSCLIP, built using a straightforward image-text contrastive learning approach with our proposed dataset. To assess the efficacy of this foundational model, we conducted primary downstream tasks including zero-shot classification, image-text retrieval, and semantic localization. When comparing RSCLIP to models not trained on the distribution of direct language descriptions, RSCLIP consistently outperformed its counterparts. Even though RSCLIP might not always surpass models trained directly with language descriptions, its performance remains highly competitive. Looking ahead, our future endeavors will explore the integration of various modalities present in remote sensing imagery, expressed in the form of language.

\ifCLASSOPTIONcaptionsoff
  \newpage
\fi

\vspace{-2mm}
\bibliographystyle{IEEEtran}
\bibliography{bibtex/IEEEexample}

\end{document}